\documentclass{article} 
\usepackage{nips15submit_e,times}

\usepackage{natbib}

\usepackage{subfigure}
\usepackage{amsmath}
\usepackage{amssymb}
\usepackage{amsfonts}
\usepackage{amsthm}
\usepackage{graphicx}
\usepackage{color}
\usepackage{url}
\usepackage{sidecap}

\DeclareMathOperator*{\argmax}{argmax}

\newcommand{\vect}[1]{\boldsymbol{#1}}

\title{A Sparse and Adaptive Prior for Time-Dependent Model Parameters}
\author{Dani Yogatama \qquad Bryan R.~Routledge \\
Carnegie Mellon University \\
\texttt{\{dyogatama@cs.,routledge@\}cmu.edu}
\And
Noah A.~Smith \\
University of Washington \\
\texttt{nasmith@cs.washington.edu} }

\nipsfinalcopy
\begin{document}

\maketitle
\vspace{-0.4cm}
\begin{abstract}
\vspace{-0.2cm}
We consider the scenario where the parameters of a probabilistic model
are expected to vary over time.  We construct a novel prior distribution that promotes sparsity and adapts the strength of correlation between parameters at successive timesteps, based on the data.  We derive approximate variational inference procedures for learning and prediction with this prior.  We test the approach on two tasks:  forecasting financial quantities from relevant text, and modeling language contingent on time-varying financial measurements.
\vspace{-0.2cm}
\end{abstract}
\vspace{-0.1cm}

\section{Introduction}

When learning from streams of data to make predictions in the future, how should we handle the timestamp associated with each instance?  Ignoring timestamps and assuming data are i.i.d.~is scalable but risks distracting a model with irrelevant ``ancient history.''  On the other hand, using only the most recent portion of the data risks overfitting to current trends and missing important time-insensitive effects.
In this paper, we seek a general approach to learning model parameters that are overall sparse, but that adapt to variation in how different effects change over time.

Our approach is a prior over parameters of an exponential family
(e.g., coefficients in  linear or logistic regression).  We assume
that parameter values shift at each timestep, with correlation between
adjacent timesteps captured using a multivariate normal distribution
whose precision matrix is restricted to a tridiagonal structure.  We
(approximately) marginalize the (co)variance parameters of this normal
distribution using a Jeffreys prior, resulting in a model that allows
smooth variation over time while encouraging overall sparsity in the
parameters.  (The parameters themselves are not given a fully Bayesian treatment.)

There are many related Bayesian approaches for time-varying model parameters \citep{belmonte,nakajima,caron},
as well as work on time-varying signal estimation \citep{angelosante,angelosante2,charles}. 
Our model has a distinctive generative story in that correlations between parameters of successive timesteps are encoded in a precision matrix.
Additionally, unlike these fully Bayesian approaches that infer full posterior distributions, we only obtain posterior mode estimates of coefficients, which has computational advantages at prediction time (straightforward MAP inference and sparsity).

We demonstrate the usefulness of our model on two tasks.
The first is a text regression problem in which an economic variable (volatility of returns) is forecast from financial reports \citep{smith2009}.  The second forecasts text by constructing a language model that conditions on highly time-dependent economic variables.

\section{Notation}
\label{sec:erm}

We assume data of the form $\{ (x_n, y_n)\}_{n=1}^N$, where each $x_n$ includes a timestamp denoted $t \in \{1, \ldots, T\}$.\footnote{In this work we assume timestamps are discretized.}  The aim is to learn a predictor that maps input $x_{N+1}$, assumed to be at timestep $T$, to output $y_{N+1}$.  In the probabilistic setting we adopt here, the prediction is MAP inference over r.v.~$Y$ given $X=x$ and a model parameterized by $\vect{\beta} \in \mathbb{R}^I$.  Learning is parameter estimation to solve:
\begin{equation}
\argmax_{\vect{\beta}} \log p(\vect{\beta}) +  
\overbrace{\sum_{n=1}^N \log \underbrace{p(y_n \mid x_n, \vect{\beta})}_{\mathrm{link}^{-1}( \vect{f}(x)^\top \vect{\beta}  )}}^{L(\vect{\beta})}  \label{eq:learn}
\end{equation}
The focus of the paper is on the prior distribution $p(\vect{\beta})$.  Throughout, we will denote the task-specific log-likelihood (second term) by $L(\vect{\beta})$ and assume 
a generalized linear model such that a feature vector function $\vect{f}$ maps inputs $x$ into $\mathbb{R}^I$ and $\vect{f}(x)^\top\vect{\beta}$ is ``linked'' to the distribution over $Y$ using, e.g., a logit or identity.  We will refer to elements of $\vect{f}$ as ``features'' and  to $\vect{\beta}$ as ``coefficients.'' 
 We assume $T$ discrete timesteps.

\section{Time-Series Prior}
\label{sec:model}
Our time-series prior draws inspiration from the probabilistic interpretation of the sparsity-inducing lasso \citep{tibshirani1996} and group lasso \citep{grouplasso}. 
In non-overlapping group lasso, features are divided into groups, and the coefficients within each group $m$  are drawn according to:
\begin{enumerate}
\item Variance $\sigma^2_m \sim$ an exponential distribution.\footnote{The exponential distribution can be replaced by the (improper) Jeffreys prior, although then the familiar Laplace distribution interpretation no longer holds \citep{figueiredo2002}.} 
\item $\vect{\beta}_m \sim \mathrm{Normal}(\vect{0}, \sigma^2_m \mathbf{I})$. 
\end{enumerate}

We seek a prior that lets each coefficient vary smoothly over time.  
A high-level intuition of our prior is that we create copies of $\vect{\beta}$, one at each timestep:  $\langle \vect{\beta}^{(1)}, \vect{\beta}^{(2)}, \ldots, \vect{\beta}^{(T)}\rangle$.  For each
feature $i$,  let the sequence $\langle \beta^{(1)}_i, \beta^{(2)}_i, \ldots, \beta^{(T)}_i \rangle$ form a group, denoted $\vect{\beta}_i$.  
Group lasso does not view coefficients in a group as explicitly correlated; they are independent given the variance parameter.  Given the sequential structure of $\vect{\beta}_i$, we replace the covariance matrix $\sigma^2\mathbf{I}$ to capture autocorrelation.  
Specifically, we assume the vector $\vect{\beta}_i$ is drawn from a multivariate normal distribution with mean zero and a $T \times T$ precision matrix $\mathbf{\Lambda}$ with the following tridiagonal form:\footnote{We suppress the subscript $i$ for this discussion; each feature $i$ has its own $\mathbf{\Lambda}_i$.}
\begin{align}
\mathbf{\Lambda} &= \frac{1}{\lambda} \mathbf{A} & = \frac{1}{\lambda} \left [ 
\begin{array}{cccccc}
1 & \alpha & 0 &  0 & \dots \\
\alpha & 1 & \alpha & 0 & \dots \\
0 & \alpha & 1 & \alpha & \dots \\
0 & 0 & \alpha & 1 & \dots \\
\vdots & \vdots & \vdots & \vdots & \ddots
\end{array}
\right ]
\label{eq:Lambda}
\end{align}
$\lambda \geq 0$ is a scalar multiplier whose role is to control sparsity in the coefficients, 
while $\alpha$ dictates the degree of correlation between coefficients in adjacent timesteps (autocorrelation).  Importantly, $\alpha$ and $\lambda$  (and hence $\mathbf{A}$ and  $\mathbf{\Lambda}$) are allowed to be different for each group $i$.

We need to ensure that  $\mathbf{A}$ is positive definite.
Fortunately, it is easy to show that for $\alpha \in (-0.5,0.5)$, the resulting $\mathbf{A}$ is positive definite.
We give a proof sketch in the supplementary materials.

\subsection{Generative Model}

Our generative model for the group of coefficients $\vect{\beta}_i = \langle \beta^{(1)}_i, \beta^{(2)}_i, \ldots, \beta^{(T)}_i \rangle$ is given by: 
\begin{enumerate}
\item $\lambda_i \sim $ an improper Jeffreys prior ($p(\lambda) \propto \lambda^{-1}$).
\item $\alpha_i \sim $ a  truncated exponential prior with parameter $\tau$.  This distribution 
forces $\alpha_i$ to fall in $(-C, 0]$, so that $\mathbf{A}_i$ is p.d. and autocorrelations are always positive (we fix $C = \frac{1}{2} - 10^{-5}$):
\begin{equation}
p(\alpha \mid \tau) = \frac{\tau \exp(-\tau(\alpha + C))\vect{1}\{-C < \alpha \leq 0 \} }{(1-\exp(-\tau C))}.
\end{equation} 
\item  $
\vect{\beta}_i \sim \mathrm{Normal}(\vect{0}, \mathbf{\Lambda}_i^{-1})$, with the precision matrix $\mathbf{\Lambda}_i$ as defined in Eq.~\ref{eq:Lambda}. 
\end{enumerate}
During estimation of $\vect{\beta}$, $\lambda_i$ and $\alpha_i$ are marginalized, giving a sparse and adaptive estimate for $\vect{\beta}$.

\subsection{Scalability}

Our design choice of the precision matrix $\mathbf{\Lambda}_i$ is driven by scalability concerns.
Instead of using, e.g., a random draw from a Wishart distribution, we specify 
the precision matrix to have a tridiagonal structure.
This induces dependencies between coefficients in adjacent timesteps
(first-order dependencies) and allows the prior to scale to
fine-grained timesteps more efficiently.
Let $N$ denote the number of training instances, $I$ the number of base features, and $T$ the number of timesteps.
A single pass of our variational algorithm (discussed in \S\ref{sec:inference}) has runtime $\mathcal{O}(I(N+T))$ and space requirement $\mathcal{O}(I(N+T))$, 
instead of $\mathcal{O}(I(N+T^2))$ for both if each $\mathbf{\Lambda}_i$ is drawn from a Wishart distribution. 
This can make a big difference for applications with large numbers of features ($I$).
Additionally, we choose the off-diagonal entries to be uniform, so we only need one $\alpha_i$ for each base feature.
This design choice restricts the expressive power of the prior but still permits flexibility in adapting to trends for different coefficients, as we will see.
The prior encourages sparsity at the group level, essentially performing feature selection:  some feature coefficients $\vect{\beta}_i$ may be driven to zero across all timesteps, while others will be allowed to vary over time, with an expectation of smooth changes.

Note that this model introduces only one hyperparameter, $\tau$, since we marginalize $\vect{\alpha} = \langle \alpha_1, \ldots, \alpha_I\rangle$ and $\vect{\lambda} = \langle \lambda_1, \ldots, \lambda_I\rangle$.

\section{Learning and Inference}
\label{sec:inference}

We marginalize $\vect{\lambda}$ and $\vect{\alpha}$ and obtain a maximum \emph{a posteriori} estimate for $\vect{\beta}$, which includes a coefficient for each base feature $i$ at each timestep $t$.
Specifically, we seek to maximize:
\begin{align}
L(\vect{\beta}) + \sum_{i=1}^I \log \int d\alpha_i \int d\lambda_i p(\vect{\beta}_i \mid \alpha_i, \lambda_i) p(\alpha_i \mid \tau) p(\lambda_i)  
\end{align}
Exact inference in this model is intractable.
We use mean-field variational inference to derive a lower bound on the above log-likelihood function.
We then apply a standard optimization technique to jointly optimize the variational parameters and the coefficients $\vect{\beta}$.
See supplementary materials for details.

\section{Experiments}
\label{sec:experiments}

We report financial forecasting experiments here and language modeling experiments in the supplementary materials. 
Each timestep in our experiments is one year.

\subsection{Forecasting Risk from Text}
\label{sec:experimentsforecasting}
In the first experiment, we apply our prior to a forecasting task.
We consider the task of predicting volatility of stock returns from financial reports of publicly-traded companies \citep{smith2009}.

In finance, \emph{volatility} refers to a measure of variation in a quantity over time; for stock returns, it is measured using the standard deviation during a fixed period (here, one year). Volatility is used
as a measure of financial risk.
Consider a linear regression model for predicting the log volatility\footnote{Similar to \citet{smith2009} and as also the standard practice in finance, we perform a log transformation, since log volatilities are typically close to normally distributed.} of a stock from a set of features (see \S{\ref{sec:dataset}} for a complete description of our features).
We can interpret a linear regression model probabilistically as drawing $y \in \mathbb{R}$ from a normal distribution with $\vect{\beta}^{\top}\vect{f}(x)$ as the mean of the normal.
Therefore, in this experiment:
$L(\vect{\beta})  =  -\sum_{t=1}^T \sum_{i=1}^{N_t} (y_{i}^{(t)} - \vect{\beta}^{(t)\top}  \vect{f}(x_{i}^{(t)}))^2$.

We apply the time-series prior to the feature coefficients $\vect{\beta}$.
When making a prediction for the test data, we use $\vect{\beta}^{(T)}$,  the set of feature coefficients for the last timestep in the training data.

\subsubsection{Dataset}
\label{sec:dataset}

We used  a collection of Securities Exchange Commission-mandated annual reports from 10,492 publicly traded companies in the U.S.
There are 27,159 reports over a period of ten years from 1996--2005 in the corpus.
These reports are known as ``Form 10-K.''
For the feature set, we downcased and tokenized the texts and selected the 101st--10,101st most frequent words as binary features. The feature set was kept the same across experiments for all models.
It is widely known in the financial community that the past history of volatility of stock returns is a good indicator of the future volatility.
Therefore, we also included the log volatility of the stocks twelve months prior to the report as a feature.
Our response variable $y$ is the log volatility of stock returns over a period of twelve months after the report is published.

\subsubsection{Results}
The year ``2002'' was used as our development data for hyperparameter tuning ($\tau$ was selected to be $1.0$).
We initialized all the feature coefficients by the coefficients from training a lasso regression on the last year of the training data (\textbf{lasso-one}).
We compare with baselines that vary in how they use training data and in how they regularize (see supplementary materials for details).
Table~\ref{tbl:results} provides a summary of experimental results. 
We report the results in mean squared error on the test set: $\frac{1}{N} \sum_{i=1}^{N} (y_i - \hat{y}_i )^2$, where $y_i$ is the true response for instance $i$ and $\hat{y}_i$ is the predicted response.

Our model consistently outperformed ridge variants, including the one with a time-series penalty (\textbf{ridge-ts}; Yogatama et al., 2011). \nocite{yogatama2011}
Note that \textbf{ridge-ts} can be obtained from our model by fixing the same $\alpha$ and $\lambda$ for all features $i$.\footnote{Specifically, our approach differs in that (i) we marginalize the hyperparameters, (ii) we allow each coefficient its own autocorrelation, and (iii) we encourage sparsity.}
Our model also outperformed the lasso variants without any time-series penalty, on average and in two out of three test sets apiece.

One of the major challenges in working with time-series data is to choose the right window size, in which the data is still relevant to current predictions. 
Our model automates this process with a Bayesian treatment of the strength of each feature coefficient's autocorrelation.
The results indicate that our model was able to learn when to trust a longer history of training data, 
and when to trust a shorter history of training data, demonstrating the adaptiveness of our prior. 

In future work, an empirical Bayesian treatment of the hyperprior $\tau$, fitting it to improve the variational bound, might lead to further improvements.

\begin{table*}[t]
\centering
\caption{
\label{tbl:results} MSE on the 10-K dataset.
For baselines, \textbf{ridge} or \textbf{lasso} indicates the regularizer used while \textbf{one} or \textbf{all} indicates the amount of training data used. \textbf{ridge-ts} is the non-adaptive time-series ridge model of \citet{yogatama2011}. See supplementary materials for details about our baselines.
The overall differences between our model and all competing models are statistically significant (Wilcoxon signed-rank test, $p < 0.01$).}
\vspace{0.1cm}
\begin{tabular}{|c||c||c|c|c|c|c|c|}
\hline
\textbf{year} & \textbf{\# examples}& \textbf{ridge-one} & \textbf{ridge-all} & \textbf{ridge-ts} &\textbf{lasso-one} & \textbf{lasso-all} & \textbf{our model}\\
\hline\hline
\textbf{2003} & 3,611 & 0.185 & 0.173 & 0.171 & \textbf{0.164} & 0.176 & 0.164\\
\textbf{2004} & 3,558 & 0.125 & 0.137 & 0.129 & 0.116 & 0.119 & \textbf{0.113}\\
\textbf{2005} & 3,474 & 0.135 & 0.133 & 0.136  & 0.124 & 0.124 & \textbf{0.122}\\
\hline\hline
\textbf{overall} & 13,488 & 0.155 & 0.154 & 0.151 & 0.141 & 0.143 & \textbf{0.139} \\  
\hline
\end{tabular}
\end{table*}
\vspace{-0.4cm}

\section{Conclusions}
\label{sec:conclusion}
We presented a time-series prior for the parameters of probabilistic models; it produces sparse models and adapts the strength of temporal effects on each coefficient separately, based on the data, without an explosion in the number of hyperparameters.
We showed how to do inference under this prior  using variational approximations.
We evaluated the prior for the task of forecasting volatility of stock returns from financial reports,
and demonstrated that it outperforms other competing models.
We also evaluated the prior for the task of modeling a collection of texts over time, i.e., predicting the probability of words given some observed real-world variables. 
We showed that the prior achieved state-of-the-art results as well.

{\small
\section*{Acknowledgments}
The authors thank anonymous reviewers for helpful feedback on
earlier drafts of this paper.
This research was supported in part by a Google research award to the
second and third authors.
This research was  supported in part by the Intelligence Advanced Research Projects
Activity  via Department of Interior National Business Center
 contract number D12PC00347.  The U.S.~Government is authorized to
 reproduce and distribute reprints for Governmental purposes
 notwithstanding any copyright annotation thereon.  The views and conclusions contained
 herein are those of the authors and should not be interpreted as
 necessarily representing the official policies or endorsements,
 either expressed or implied, of IARPA, DoI/NBC, or the U.S.~Government.
}

\bibliographystyle{icml2013}
\bibliography{paper.bib}

\end{document}


\maketitle

\section{Proof of Positive Definiteness of $\mathbf{A}$}
We show that for $\alpha \in (-0.5,0.5)$, the covariance matrix $\mathbf{A}$ used by our time-series prior is always positive definite.
\begin{proof}[Proof sketch]
Since $\mathbf{A}$ is a symmetric matrix,  we verify that each of its principal minors have strictly positive determinants.
The principal minors of $\mathbf{A}$ are uniform tridiagonal symmetric matrices,
and the determinant of a uniform tridiagonal $N\times N$ matrix can be written as
$\prod_{n=1}^{N} \left\{ 1 + 2 \alpha \cos\left(  \frac{(n+1)\pi}{N+1} \right) \right\}$ (see, e.g., \citet{foxes} for the proof).
Since $\cos(x) \in [-1,1]$, if $\alpha \in (-0.5,0.5)$, the determinant is always positive.
Therefore, $\mathbf{A}$ is always p.d.~for $\alpha \in (-0.5,0.5)$.
\end{proof}

\section{Details of Learning and Inference}

Recall that during learning we marginalize $\vect{\lambda}$ and $\vect{\alpha}$ and obtain a maximum \emph{a posteriori} estimate for $\vect{\beta}$, which includes a coefficient for each base feature $i$ at each timestep $t$.
The objective function that we seek to maximize is:
\begin{align}
L(\vect{\beta}) + \sum_{i=1}^I \log \int d\alpha_i \int d\lambda_i p(\vect{\beta}_i \mid \alpha_i, \lambda_i) p(\alpha_i \mid \tau) p(\lambda_i)  
\end{align}
Unfortunately, exact inference in this model is intractable, so we use mean-field variational inference to derive a lower bound on the above log-likelihood function.

We introduce fully factored variational distributions for each $\lambda_i$ and $\alpha_i$.
For $\lambda_i$, we use a Gamma distribution with parameters $a_i, b_i$ as our variational distribution:
\begin{align*}
q_i(\lambda_i \mid a_i, b_i) = \frac{ \lambda_i^{a_i-1} \exp(-\lambda_i/b_i)} {b_i^{a_i} \Gamma(a_i)}
\end{align*}
Therefore, we have $\mathbb{E}_{q_i}[\lambda_i] = a_i b_i$, $\mathbb{E}_{q_i}[\lambda_{i}^{-1}] = ((a_i-1)b_i)^{-1}$, and $\mathbb{E}_{q_i}[\log \lambda_{i}] = \Psi(a_i) + \log b_i$ ($\Psi$ is the digamma function).

For $\alpha_i$, we choose the form of our variational distribution to be the same truncated exponential distribution as its prior, with parameter $\kappa_i$, denoting this distribution $q_i(\alpha_i \mid \kappa_i)$.
We have 
\begin{align}
 \mathbb{E}_{q_i}[\alpha_i] &= \int^{0}_{-C} \alpha_i \frac{\kappa_i \exp(-\kappa_i(\alpha_i+C)) }{1-\exp(-\kappa_i C)} d\alpha_i \nonumber \\
&= \frac{1}{\kappa_i}-\frac{C}{1-\exp(- \kappa_i C)}
\end{align}
We let $q$ denote the set of all variational distributions over $\vect{\lambda}$ and $\vect{\alpha}$.

The variational bound $B$ that we seek to maximize is given in Figure~\ref{fig:varbound}.
Our learning algorithm involves optimizing with respect to variational parameters $\vect{a}$, $\vect{b}$, and $\vect{\kappa}$, and the coefficients $\vect{\beta}$.
We employ the L-BFGS quasi-Newton method \citep{liu1989}, for which we need to compute
the gradient of $B$.  We turn next to each part of this gradient.

\begin{figure*}
\small
\begin{eqnarray*}
B(\vect{a}, \vect{b}, \vect{\kappa}, \vect{\beta})  
&\propto&  L(\vect{\beta}) + \sum_{i =1}^I \left\{  \frac{1}{2} (-T \mathbb{E}_q[\log \lambda_{i}]  \fbox{$-\mathbb{E}_q[\log \det{\mathbf{A}_i^{-1}}]$} ) - \mathbb{E}_q[\lambda_{i}^{-1}]\frac{1}{2} \vect{\beta}_{i}^{\top}\mathbb{E}_q[\mathbf{A}_i]\vect{\beta}_{i} \right\} \\
&&  + \sum_{i=1}^I \left\{ - (\mathbb{E}_q[\alpha_{i}]+C)\tau - \mathbb{E}_q[\log \lambda_{i}] \right\} - \sum_{i =1}^I \left\{ (a_{i} - 1) \mathbb{E}_q[\log \lambda_{i}] - \frac{\mathbb{E}_q[\lambda_{i}]}{b_{i}} - \log \Gamma({a_{i}}) - a_{i} \log b_{i} \right\}  \\
&&  - \sum_{i =1}^I \left\{ \log \kappa_{i} -\kappa_{i}(\mathbb{E}_q[\alpha_{i}]+C) - \log (1 - \exp(-\kappa_{i}C)) \right\}
\end{eqnarray*}
\caption{\label{fig:varbound}
The boxed expression is further bounded by $- \log \det \mathbb{E}_q[\mathbf{A}_i]$  using Jensen's inequality, giving a new lower bound we denote by $B'$.}
\end{figure*}

\subsection{Coefficients $\vect{\beta}$}

For $ 1 < t < T$, the first derivative with respect to time-specific coefficient $\beta_{i}^{(t)}$ is:
\begin{equation}
 \frac{\partial B}{\partial\beta_{i}^{(t)}} = \frac{\partial L}{\partial \beta_{i}^{(t)}}  - \frac{1}{2}\mathbb{E}[\lambda_i^{-1}] \left( \mathbb{E}[\alpha_{i}](\beta_{i}^{(t-1)}+\beta_{i}^{(t+1)}) + 2\beta_{i}^{(t)} \right)
\end{equation}
We can interpret the first derivative as including a penalty scaled by $\mathbb{E}[\lambda_{i}^{-1}]$.
We rewrite this penalty as:
\begin{align*}
\mathbb{E}[\lambda_i^{-1}] \left( \vphantom{1 - \mathbb{E}[\alpha_{i}] ) 2\beta_{i}^{(t)} } \right. & (1 - \mathbb{E}[\alpha_{i}] ) & & \cdot 2\beta_{i}^{(t)} \\
& + \mathbb{E}[\alpha_{i}] && \cdot (\beta_{i}^{(t)} - \beta^{(t-1)}_{i}) \\
& + \mathbb{E}[\alpha_{i}] & &\left. \cdot (\beta_{i}^{(t)}  - \beta^{(t+1)}_{i}) \right)
\end{align*}
This form makes it clear that the penalty depends on
$\beta^{(t-1)}_{i}$ and $\beta^{(t+1)}_{i}$, penalizing the difference between $\beta^{(t)}_i$ and these time-adjacent coefficients proportional to $\mathbb{E}[\alpha_{i}]$.

The form bears strong similarity to the first derivative of the time-series (log-)prior introduced in \citet{yogatama2011}, which depends on fixed, global hyperparameters analogous to  our $\alpha$ and $\lambda$.  Because our approach does not require us to specify scalars playing the roles of ``$\mathbb{E}[\lambda_i^{-1}]$'' and ``$\mathbb{E}[\alpha_{i}]$'' in advance,
it is possible for each feature to have its own autocorrelation.
Obtaining the same effect in their model would require careful tuning of $\mathcal{O}(I)$ hyperparameters, which is not practical.

It also has some similarities to the fused lasso penalty \citep{tibshirani-fused}, which is intended to encourage sparsity in the differences between features coefficients across timesteps.
Our prior, on the other hand, encourages smoothness in the differences, with additional sparsity at the feature level.

\subsection{Variational Parameters for $\vect{\alpha}$ and $\vect{\lambda}$}

Recall that the variational distribution for $\lambda_i$ is a Gamma distribution with parameters $a_i$ and $b_i$.

\paragraph{Precision matrix scalar $\vect{\lambda}$.} The first derivative for variational parameters $\vect{a}$ is easy to compute: 
\begin{equation}
\frac{\partial B}{\partial a_i} = \left(-\frac{T}{2} - a_{i} \right)\Psi_1(a_i) + \frac{\vect{\beta}_{i}^{\top}\mathbb{E}[\mathbf{A}_i]\vect{\beta}_{i}}{2  b_{i}(a_{i}-1)^{2}} + 1
\end{equation}
where $\Psi_1$ is the trigamma function. 
We can solve for $\vect{b}$ in closed form given the other free variables:
\begin{equation}
b_{i} = \frac{\vect{\beta}_{i}^{\top}\mathbb{E}[\mathbf{A}_i]\vect{\beta}_{i} }{  (a_{i}-1)T}
\end{equation}
We therefore treat $\vect{b}$ as a function of $\vect{a}$, $\vect{\kappa}$, and $\vect{\beta}$ in optimization.

\paragraph{Off-diagonal entries $\vect{\alpha}$.}
First, notice that using Jensen's inequality: $\mathbb{E}[\log \det{\mathbf{A}_i^{-1}}] = \mathbb{E}[- \log \det{\mathbf{A}_i}] \geq - \log \det \mathbb{E}[\mathbf{A}_i]$ due to the fact that $- \log \det \mathbf{A}_i$ is a convex function. 
Furthermore, for a uniform symmetric tridiagonal matrix like $\mathbf{A}_i$, the log determinant can be computed in closed form as follows \citep{foxes}:
\begin{align*}
\log \det \mathbb{E}[\mathbf{A}_i] =& \log \left( \prod_{t=1}^{T} 1 + 2 \mathbb{E}[\alpha_i]  \cos\left(  \frac{(t+1)\pi}{T+1} \right) \right) \\
=& \sum_{t=1}^{T} \log \left( 1 + 2 \mathbb{E}[\alpha_i]  \cos\left( \frac{(t+1)\pi}{T+1} \right) \right)
\end{align*}
We therefore maximize a lower bound on $B$, making use of the above to calculate first derivatives with
 respect to $\kappa_i$:
\begin{align*}
\frac{\partial B'}{\partial\kappa_i} =& -\tau \frac{\partial\mathbb{E}[\alpha_{i}]}{\partial\kappa_i} -\frac{1}{\kappa_{i}} + C + \mathbb{E}[\alpha_{i}] + \frac{\partial \mathbb{E}[\alpha_{i}]}{\partial \kappa_i}\kappa_{i} \\
&+ \frac{C \exp(-C\kappa_{i})}{1-\exp(-C\kappa_{i})} + \frac{1}{2} \frac{ \partial \log \det \mathbb{E}[\mathbf{A}_i]}{\partial \kappa_i} \\
&- \frac{1}{2}\mathbb{E}[\lambda_{i}^{-1}] \frac{\partial \vect{\beta}_{i}^{\top}\mathbb{E}[\mathbf{A}_i]\vect{\beta}_{i} }{\partial \kappa_i}
\end{align*}
The partial derivatives $\frac{\partial \mathbb{E}[\alpha_{i}]}{\partial \kappa_i}$, $\frac{\partial \log \det \mathbb{E}[\mathbf{A}_i]}{\partial \kappa_i}$, and $\frac{\partial \vect{\beta}_{i}^{\top}\mathbb{E}[\mathbf{A}_i]\vect{\beta}_{i}}{\partial \kappa_i} $ are easy to compute. We omit them for space.

\subsection{Implementation Details}
A well-known property of numerical optimizers like the one we use (L-BFGS; \citet{liu1989}) is the failure to reach optimal values exactly at zero.  
Although theoretically strongly sparse, our prior only produces weak sparsity in practice.
Future work might consider a more principled proximal-gradient algorithm to obtain strong sparsity \citep{proxgrad1,proxgrad2,proxgrad3}.

If we expect feature coefficients at specific timesteps to be sparse as well, it is straightforward to incorporate additional terms in the objective function that encode this prior belief (analogous to an extension from group lasso to \emph{sparse} group lasso).
For the tasks we consider in our experiments, we found that it does not substantially improve the overall performance.
Therefore, we keep the simpler bound given in Figure~\ref{fig:varbound}. 

\section{Baselines} 
\label{sec:baselines}
We compare our approach to a range of baselines.
At each test year, we only used training examples that come from earlier years.  Our baselines vary in how they use this earlier data and in how they regularize.
\begin{itemize}
\item \textbf{ridge-one}:  ridge regression \citep{hoerl1970}, trained on only examples from the year prior to the test data (e.g., for the 2002 task, train on examples from 2001)
\item \textbf{ridge-all}:  ridge regression trained on the full set of past examples (e.g., for the 2002 task, train on examples from 1996--2001)
\item \textbf{ridge-ts}:  the non-adaptive time-series ridge model of \citet{yogatama2011}
\item \textbf{lasso-one}:  lasso regression \citep{tibshirani1996}, trained on only examples from the year prior to the test data\footnote{Brendan O'Connor (personal communication) has established the superiority of the lasso to the support vector regression method of \citet{smith2009} on this dataset; lasso is a strong baseline for this problem.}
\item \textbf{lasso-all}:  lasso regression trained on the full set of past examples
\end{itemize}
In all cases, we tuned hyperparameters on a development data. 
Note that, of the above baselines, only \textbf{ridge-ts} replicates the coefficients at different timesteps (i.e., $IT$ parameters); the others have only $I$ time-insensitive coefficients.

The model with our prior always uses all training examples that are available up to the test year (this is equivalent to a sliding window of size infinity).  Like \textbf{ridge-ts}, our model trusts more recent data more, allowing coefficients farther in the past to drift farther away from those most relevant for prediction at time $T + 1$.  Our model, however, adapts the ``drift'' of each coefficient separately rather than setting a global hyperparameter.

\section{More Experiments: Text Modeling in Context}
\label{sec:experimentsmodeling}
In this experiment, we consider a hard task of modeling a collection of texts over time conditioned on economic measurements.
The goal is to predict the probability of words appearing in a document, based on the ``state of the world'' at the time the document was authored.
Given a set of macroeconomic variables in the U.S.~(e.g., unemployment rate, inflation rate, average housing prices, etc.), we want to predict what kind of texts will be produced at a specific timestep.
These documents can be written by either the government or publicly-traded companies directly or indirectly affected by the current economic situation. 

\subsection{Model}

Our text model is a sparse additive generative model (SAGE; \citet{sage}).
In SAGE, there is a background lexical distribution that is perturbed additively in the log-space.
When the effects are due to a (sole) feature $f(x)$, the probability of a word is: 
\begin{align*}
p(w \mid \vect{\theta},\vect{\beta},x) =  \frac{\exp (\theta_w + \beta_{w}f(x)) }{\sum_{w' \in V} \exp (\theta_{w'} + \beta_{w'}f(x))}
\end{align*}
where $V$ is the vocabulary, $\vect{\theta}$ (always observed) is the vector of background log-frequencies of words in the corpus, $f(x)$ (observed) is the feature derived from the context $x$, and $\vect{\beta}$ is the feature-specific deviation.

Notice that the formulation above is easily extended to multiple effects with coefficients $\vect{\beta}$.
In our experiment, we have 117 effects (features), each with its own $\vect{\beta}_i$.
The first 50 correspond to U.S.~states, plus an additional feature for the entire U.S., and they are observed for each text since each text is associated with a known set of states (discussed below).
We assume that texts that are generated in different states have distinct characteristics;
for each state, we have a binary indicator feature. 
The other 66 features depend on observed macroeconomics variables at each timestep (e.g., unemployment rate, inflation rate, house price index, etc.).
Given an economic state of the world, we hypothesize that there are certain words that are more likely to be used, and each economic variable has its own (sparse) deviation from the background word frequencies.
The generative story for a word at timestep $t$ associated with (observed) features $\vect{f}(x^{(t)})$ is:
\begin{itemize}
\item Given observed real-world observed variables $x^{(t)}$, draw  word $w$ from a multinomial distribution $p(w \mid \vect{\theta}^{(t)}, \vect{\beta}^{(t)}, x^{(t)}) \propto \exp (\theta_w^{(t)} + \vect{\beta}_w^{(t)\top} \vect{f}(x^{(t)}) )$.
\end{itemize}

Our $L(\vect{\beta})$ is simply the negative log-loss function commonly used in multiclass logistic regression:
$L(\vect{\beta}) = \sum_{t=1}^T \sum_{i=1}^{N_t} \log p(\vect{w}_i^{(t)} \mid \vect{\theta}^{(t)}, \vect{\beta}^{(t)}, x^{(t)}_i)$.
We apply our time-series prior from \S\ref{sec:model} to $\vect{\beta}$.
$\vect{\theta}^{(t)}$ is fixed to be the log frequencies of words at timestep $t$.  
For a single feature, coefficients over time for different classes (words) are assumed to be generated from the same prior.

\subsection{Dataset}
\label{sec:dataset2}
There is a great deal of text that is produced to describe current macroeconomic events.  We conjecture that the connection between the economy and the text will have temporal dependencies (e.g., the amount of discussion about housing or oil prices might vary over time).  We use three sources of text commentary on the economy.  The first is a subset
of the 10-K reports we used in our risk forecasting experiment.  
We selected the 10-K reports of 200 companies chosen randomly from the top quintile of size (measured by beginning-of-sample market capitalization).
This gives us a manageable sample of the largest U.S. companies.
Each report is associated with the state in which the company's head office is located.
Our next two data sources come from the Federal Reserve System, the primary body responsible for monetary policy in the U.S.\footnote{
For an overview of the Federal Reserve System, see the Federal Reserve's ``Purpose and Functions'' document at \url{http://www.federalreserve.gov/pf/pf.htm}.}  The Federal Open Market Committee (FOMC) meets roughly eight times per year to discuss economic conditions and set monetary policy.  Prior to each meeting, each of the twelve regional banks write an informal ``anecdotal'' description of economic activity in their region as well as a national summary.  This ``Beige Book''  is akin to a blog of economic activity released prior to each meeting.  Each FOMC meeting also produces a transcript of the discussion.  For our experiments here, we focus on text from 1996--2006.\footnote{All the text is freely available at \url{http://www.federalreserve.gov}. The
Beige Book  is released to the public prior to each meeting.  The transcripts are released five years after the meetings.} As a result, we have 2,075 documents in the final corpus, consisting of 842 documents of the 10-K reports, 89 documents of the FOMC meeting transcripts, and 1,144 documents of the Beige Book summaries.  

We use the 501st--5,501st most frequent words in the dataset.
We associated the FOMC meeting transcripts with all states.
The ``Beige Book'' texts were produced by the Federal Reserve Banks. 
There are twelve Federal Reserve Banks in the United States, each serving a collection of states. 
We associated texts from a Federal Reserve Bank with the states that it serves. 

\begin{table*}[t]
\centering
\caption{\label{tbl:results2} Negative log-likelihood of the documents on various test sets (lower is better).
The first test year (2003) was used as our development data. 
Our model uses  the sparse adaptive prior in~\S\ref{sec:model}.} 
\begin{tabular}{|c||c||c|c|c|c|c|c|}
\hline
& \textbf{\# tokens} & \textbf{ridge-one} & \textbf{ridge-all} & \textbf{ridge-ts} & \textbf{lasso-one}& \textbf{lasso-all}  &\textbf{our model} \\ 
\textbf{year} & ($\times 10^6$) & ($\times 10^3$) & ($\times 10^3$) & ($\times 10^3$) & ($\times 10^3$) & ($\times 10^3$) & ($\times 10^3$) \\
\hline\hline
\textbf{2004}&1.5 & 2,975 & 3,004 & 2,975 & 2,975 & 3,004 & \textbf{2,974} \\
\textbf{2005}&1.9 & 2,999 & 3,027 & 2,997 & 2,998 & 3,027 & \textbf{2,997} \\
\textbf{2006}&2.3 & 2,916 & 2,922 & 2,913 & 2,912 & 2,922 & \textbf{2,912} \\ \hline\hline
\textbf{overall} & 6.8 & 11,626 & 11,718 & 11,619 & 11,620 & 11,718 & \textbf{11,618}\\
\hline
\end{tabular}
\end{table*}

Quantitative U.S.~macroeconomic data was obtained from the Federal Reserve Bank of St.~Louis data repository (``FRED'').  We used standard measures of economic activity focusing on output (GDP), employment, and specific markets (e.g., housing).\footnote{For growing output series, like GDP,
we calculate growth rates as log differences.} 
We use equity market returns for the U.S.~market as a whole and various industry and characteristic portfolios.\footnote{Returns are monthly, excess of the risk-free rate, and continuously compounded.  The data are  from CRSP and are available for these portfolios at
\url{http://mba.tuck.dartmouth.edu/pages/faculty/ken.french/data_library.html}.}
They are used as $\vect{f}(x)$ in our model; in addition to state indicator variables, there are 66 macroeconomic variables in total.

\subsection{Results}
\label{sec:experimentsmodelingresults}

We score models by computing the negative log-likelihood on the test dataset:\footnote{Out-of-vocabulary items are ignored.}
$- \sum_{i=1}^{N} \log p(\vect{w}_i^{(T + 1)} \mid \vect{\theta}^{(T)}, \vect{\beta}^{(T)}, x^{(T + 1)}_i)$.
We initialized all the feature coefficients by the coefficients by training a lasso regression on the last year of the training data (\textbf{lasso-one}).
The first test year (i.e., 2003) was used as our development data for hyperparameter tuning ($\tau$ was selected to be $.001$).
Table~\ref{tbl:results2} shows the results for the six models we compared.
Similar to the forecasting experiments, at each test year, we trained only on documents from earlier years.
When we collapsed all the training data and ignored the temporal dimension (ridge-all and lasso-all), the background log-frequencies $\vect{\theta}^{(t)}$ are computed using the entire training data, which is different compared to the background log-frequencies for only the last timestep of the training data. 
Our model outperformed all ridge and lasso variants, including the one with a time-series penalty \citep{yogatama2011}, in terms of negative log-likelihood on unseen dataset.

\bibliographystyle{icml2013}
\bibliography{paper.bib}